\documentclass[10pt,conference]{IEEEtran}
\usepackage{times}
\usepackage{graphicx}
\usepackage{subcaption}
\usepackage{amsmath}
\usepackage{amssymb}
\usepackage{amsfonts}
\usepackage{comment}
\usepackage{color}
\usepackage{multicol}
\usepackage{booktabs}
\usepackage[numbers]{natbib}
\usepackage{notoccite}
\usepackage[bookmarks=true]{hyperref}
\usepackage[font=small,labelfont=bf]{caption}

\usepackage[top=1in, left=0.75in, bottom=0.75in, right=0.75in]{geometry}

\IEEEoverridecommandlockouts
\begin{document}
\title{A Multi-Domain Feature Learning Method for Visual Place Recognition}

\author{Peng Yin$^{1,\mathbf{*}}$, Lingyun Xu$^1$, Xueqian Li$^3$, Chen Yin$^{4}$, Yingli Li$^1$,\\ Rangaprasad Arun Srivatsan$^3$, Lu Li$^3$, Jianmin Ji$^{2,\mathbf{*}}$, Yuqing He$^1$
\thanks{
This paper was supported by the National Natural Science Foundation of China (No. 61573386, No. 91748130, U1608253) and Guangdong Province Science and Technology Plan projects (No. 2017B010110011).

P. Yin, L. Xu, Y. Li and Y. He are with the State Key Laboratory of Robotics, Shenyang Institute of Automation, Chinese Academy of Sciences, Shenyang, University of Chinese Academy of Sciences, Beijing. 
{(yinpeng, xulingyun, liyingli, heyuqing@sia.cn)}
J. Ji is with the School of Computer Science and Technology, University of Science and Technology of China, Hefei Anhui.
{(jianmin@ustc.edu.cn)}
X. Li, R.A. Srivatsan and L. Li are with the Biorobotics Lab, Robotics Institute, Carnegie Mellon University, Pittsburgh, PA 15213, USA.
{(xueqianl, arangapr, lilu12@andrew.cmu.edu)}
Y. Chen is with the School of Computer Science, University of Beijing University of Posts and Telecommunications, Beijing.
{(chenyin@bupt.edu.cn)}

{(Corresponding author: Peng Yin, Jianmin Ji)}
}
}

\maketitle
\begin{abstract}
Visual Place Recognition (VPR) is an important component in both computer vision and robotics applications, thanks to its ability to determine whether a place has been visited and where specifically.
A major challenge in VPR is to handle changes of environmental conditions including weather, season and illumination.
Most VPR methods try to improve the place recognition performance by ignoring the environmental factors, leading to decreased accuracy decreases when environmental conditions change significantly, such as day versus night.
To this end, we propose an end-to-end conditional visual place recognition method. 
Specifically, we introduce the multi-domain feature learning method (MDFL) to capture multiple attribute-descriptions for a given place, and then use a feature detaching module to separate the environmental condition-related features from those that are not. The only label required within this feature learning pipeline is the environmental condition.
Evaluation of the proposed method is conducted on the multi-season \textit{NORDLAND} dataset, and the multi-weather  \textit{GTAV} dataset. 
Experimental results show that our method improves the feature robustness against variant environmental conditions.
\end{abstract}

\section{Introduction}
In the last decade, the robotics community has achieved numerous breakthroughs in vision-based simultaneous localization and mapping (SLAM)~\cite{SLAM:vslam2} that have enhanced the navigation abilities of unmanned ground vehicles (UGV) and unmanned aerial vehicles (UAV) in complex environment.
Visual place recognition (VPR)~\cite{VPR:survey} or loop closure detection (LCD) helps robots to find loop closure in SLAM framework and is an essential element for accurate mapping and localization.
Although many methods
have been proposed in recent years, 
VPR is still a challenging problem under varying environmental conditions.
Traditional VPR approaches that use handcrafted features to learn place descriptors for local scene description, often fail to extract valid features when encountering significant changes~\cite{VPR:SeqSLAM} in environmental conditions, such as changes in season, weather, illumination, as well as viewpoints.

\begin{figure}
	\centering
	\includegraphics[width=\linewidth]{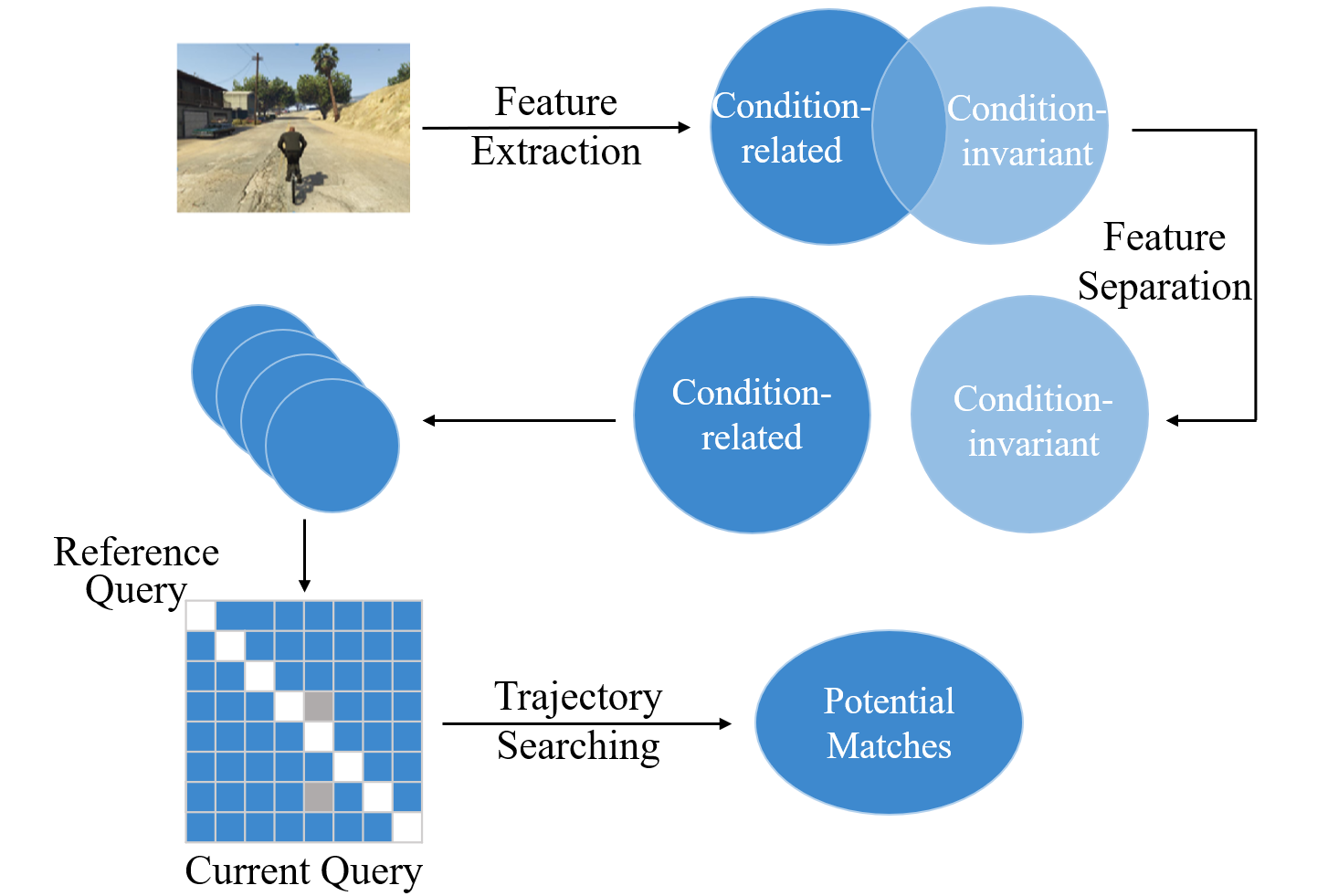}
	\caption{The pipeline of our proposed conditional visual place recognition method. 
        In summary, there exists three core modules:
        1) a CapsuleNet~\cite{CNN:CapsuleNet} based feature extraction module that is responsible for extracting condition-related and condition-invariant features from the raw visual inputs; 
        2) a condition enhanced feature separation module to further separate condition-related ones in the joint feature-distribution; 
        3) a trajectory searching mechanism for finding best matches based on the feature differences of query trajectory features.}
	\label{fig:pipeline}
\end{figure}

Ideally, the place recognition method should be able to capture condition-invariant features for robust loop closure detection, since the appearance of scene objects (e.g., roads, terrains and houses) is often highly related to environmental conditions, and that each object has its own appearance distribution under variant conditions.
To the best of our knowledge, there are few VPR methods that have explored how to improve the place recognition performance against variant environmental conditions~\cite{VPR:conditionSeqSLAM}. A major drawback of these methods is that the change in environmental conditions affects the local features, resulting in decreased accuracy of VPR.
In this paper, we propose the condition-directed visual place recognition method to address this issue.
Our work consists of two parts: feature extraction and feature separation.



Firstly, in the feature extraction step, we utilize a CapsuleNet-based network~\cite{CNN:CapsuleNet} to extract multi-domain place features, as shown in Fig.~\ref{fig:pipeline}.
Traditional convolutional neural network (CNN) is efficient in object detection
, regression
and segmentation
, but as pointed out by Hinton, the inner connections of objects are easily lost with the deep convolutional and max pooling operations.
For instance, in face detection tasks, even if the facial objects (nose, eyes, mouth, lips) are in incorrect layouts, the traditional CNN method may still consider the image as a human face, since it contains all the necessary features of a human face.
This problem also exists in place recognition tasks, since different places may contain similar objects but with different arrangements. 
CapsuleNet uses an dynamic routing method to cluster the shallow convolutional layer features in an unsupervised way.
In this paper, we demonstrate another application of CapsuleNet, which could capture feature distribution under specific conditions.

The main contributions of this work can be summarized as follows:
\begin{itemize}
	\item We propose the use of CapsuleNet-based feature extraction module, and show its robustness in the conditional feature learning for the visual place recognition task.
	\item We propose a feature separation method for the visual place recognition task, where features are indirectly separated based on the relationship between condition-related and condition-invariant features in an information-theoretic view.
\end{itemize}

The outline of the paper is as follows: Section~\ref{sec:related_works} introduces the related works on visual-based place recognition methods. 
Section~\ref{sec:proposed_method} describes our conditional visual place recognition method, which has two components: feature extraction and feature separation. 
In Section~\ref{sec:experiment_results}, 
we evaluate the proposed method on two challenging datasets: the \textit{NORDLAND}~\cite{DATASET:KITTI} dataset which has same trajectories under multiple season conditions and a \textit{GTAV} dataset which is generated on the same trajectory under different weather conditions in a game simulator.
Finally, we provide concluding remarks in Section~\ref{sec:conclusion}.

\section{Related Work}
\label{sec:related_works}
Visual place recognition (VPR) methods have been well studied in past several years, and can be classified into two categories: feature- and appearance-based.
In feature-based VPR, descriptive features are transformed into local place descriptors.
Then, place recognition can be achieved by extracting the current place descriptors and searching similar place indexes in the bag of words.
On the contrary, appearance-based VPR uses feature descriptors that are extracted from the entire image, and performs place recognition by assessing feature similarities.
SeqSLAM~\cite{VPR:SeqSLAM} describes image similarities by directly using the sum of absolute difference (SAD) between frames, while  vector of locally aggregated descriptors (VLAD)~\cite{FeatureCapturer:VLAD} aggregates local invariant features into a single feature vector and uses Euclidean distance between vectors to quantify image similarities.

Recently, many works have investigated CNN-based features for appearance-based VPR tasks. 
S\"{u}nderhauf \emph{et al}.~\cite{VPR:cnnVPR} first used pre-trained VGG model to extract middle-layer CNN outputs as image descriptors in the sequence matching pipeline. 
However, a pre-trained network can not be further trained for place recognition task, since the data labels are hard to define in VPR task. 
Recently, Chen \emph{et al}.~\cite{VPR:cnnSeqSLAM} and Garg \emph{et al}.~\cite{VPR:conditionSeqSLAM} address the conditional invariant VPR as an image classification task and rely on precise but expensive human labeling for semantic labels. 
Arandjelovic \emph{et al}.~\cite{FeatureCapture:NetVLAD} developed NetVLAD, which is a modified form of the VLAD features, with CNN networks to improve the feature robustness. 
 
The approach that comes closest to our method is the work of Porav~\emph{et al.}~\cite{VPR:Adversarial_VPR}, 
where they learn invertible generators based on the CycleGAN \cite{CNN:CycleGAN}, 
The original CycleGAN method can transform the image from one domain to another domain, but such transformation is limited to only two domains. 
Thus, for multiple domain place recognition task, the method of Porav~\emph{et al.} requires transformation model between each pair of conditions. In contrast, our method can learn more than two conditions in the same structure.

\section{Proposed Method}
\label{sec:proposed_method}
In this section, we investigate the details of two core modules in our conditional visual place recognition method.
\subsection{Feature Extraction}
\label{sec:feature_extraction}

\subsubsection{VLAD}

VLAD is a feature encoding and pooling method, which encodes a set of local feature descriptors extracted from an image by using a clustering method such as K-means clustering. 
For the feature extraction module, we extract multi-domain place features from the raw image, by utilizing a CapsuleNet module.
Let $q_{ik}$ be the strength of the association of data vector $x_{i}$ to the cluster $\mu_{k}$, such that $q_{ik}\geq0$ and $\sum_{k=1}^{K}q_{ik}=1$, where $K$ is the clusters number. 
VLAD encodes feature $x$ by considering the residuals
\begin{align}
v_{k}=\sum_{i=1}^{N}q_{ik}(x_{i}-\mu_{k}),
\label{ep:VLAD}
\end{align}
and the joint feature description $\{v_{1}, v_{2}, v_{3},..., v_{N}\}$, where $N$ is the local features number.

Assume we can extract $N$ lower feature descriptors (each is denoted as $x_{i}$) from the raw image, we can construct a new VLAD like module with the following equation,
\begin{align}
v_{k}=\sum_{i=1}^{N}Q_{k}(l_{i})r(x_{i}, \mu_{k}),
\label{ep:Caps}
\end{align}
where $r(x_{i}, \mu_{k})$ is the residual function measuring similarities between $x_{i}$ and $\mu_{k}$, and $Q_{k}(l_{i})$ is the weighting of capsule vector $l_{i}$ involved with the $k^{\text{th}}$ cluster center.


\subsubsection{Modified CapsuleNet}
In order to transform Eq.\ref{ep:Caps} into an end-to-end learning block, we consider two aspects: 
\begin{enumerate}
	\item Constructing the residual function $r(x_{i}, \mu_{k})$;
	\item Assigning the weights $Q_{k}(l_{i})$.
\end{enumerate}
With lower layer features extracted from the shallow convolution layer, 
we use $N\times D_{property}$ matrix to map lower level features into higher level features, where $N$ is the CNN unit number in the shallow convolution layer. 

If we want to integrate the lower-higher feature mapping within a single layer, the local lower level feature $x_{i}$ should have 
a linear mapping layer to represent the residual function $Q_{i}$
\begin{align}
 f(x_{i}, \mu_{k})&=\mathbf{W}_{ik}x_{i} + b_{k} \nonumber \\
 r(x_{i},\mu_{k})&= f(x_{i}, \mu_{k}) - \frac{1}{N}\sum_{k} f(x_{i}, \mu_{k}),
\label{ep:linear_mapping}
\end{align}
where $\mathbf{W}_{ik}$ and $b_{k}$ are the linear transformation weighting and bias for the $k^{\text{th}}$ capsule center. 

Furthermore, to estimate the local capsule features weighting $Q_{k}(x_{i})$, we apply a soft assignment estimation defined as
\begin{align}
Q_{k}(x_{i})=\frac{\exp(b_{ik})}{\sum_{j}^{K}\exp(b_{ij})},
\label{ep:Q_weighting}
\end{align}
where $b_{ik}$ is the probability that the $i^{\text{th}}$  local capsule feature belonging to $k^{\text{th}}$  capsule cluster $c_{k}$. Therefore, Eq.\ref{ep:Caps} can be written in the following format,
\begin{align}
v_{k}=\sum_{i=1}^{N}\frac{\exp(b_{ik})}{\sum_{j}^{K}\exp(b_{ij})}(f(x_{i}, \mu_{k}) - \frac{1}{N}\sum_{k} f(x_{i}, \mu_{k})).
\label{ep:feature_extraction}
\end{align}

In order to learn the parameters $b_{ij}, {W}_{ik}$, and $ c_{k}$, we apply the iterative dynamic routing mechanism as described in~\cite{CNN:CapsuleNet}.
For the output of $N_{object}$ higher level features, we assume the last $D_{C}$ dimensions are assigned as the condition features, e.g. $D_{C}=4$ is in the case where the condition is \textit{season}.

\subsection{Feature Separation}
\begin{figure}
	\centering
	\includegraphics[width=0.4\linewidth]{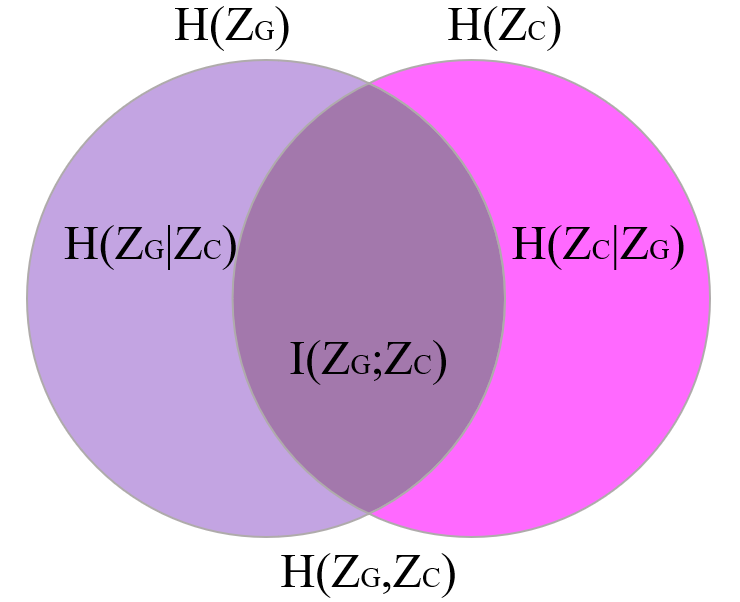}
	\caption{The relationship of condition-related $Z_{C}$ and condition-invariant $Z_{G}$ feature in the information theory view.}
	\label{fig:information}
\end{figure}

In the previous section, we described the feature extraction module $p_{\theta}$. In this section, we use an additional decoder module $q_{\phi}$, and two reconstruction modules on feature $\mathcal{L}_{Feature}$ and image $\mathcal{L}_{Image}$ domain to achieve the feature separation. 
Naturally, condition-invariant feature $Z_{G}$ and condition-related feature $Z_{C}$ are highly correlated.
Fig.~\ref{fig:information} shows the relationship between information $Z_{G}$ and $Z_{C}$. $H(Z_{G}, Z_{C})$ and $I(Z_{G};Z_{C})$ are the joint entropy and the mutual entropy respectively, while $H(z_{G}|z_{C})$ and $H(z_{C}|z_{G})$ are the conditional entropy. 
From the view of information theory, feature separation can be achieved in the following ways:
\begin{itemize}
	\item Decrease the conditional entropy $H(z|x)$: less conditional entropy enforces the unique mapping from $x\in \chi$ to $z\in (Z_{G}, Z_{C})$;
	\item Improve the geometric feature extraction capability: the more accurate geometry we capture, the higher LCD accuracy we can achieve;
	\item Reduce the mutual entropy $I(Z_{G};Z_{C}|x)$: use environmental conditions to direct feature extraction.
\end{itemize}
We add these three restrictions in our feature separation module.

\begin{figure*}
	\centering
	\includegraphics[width=0.6\linewidth]{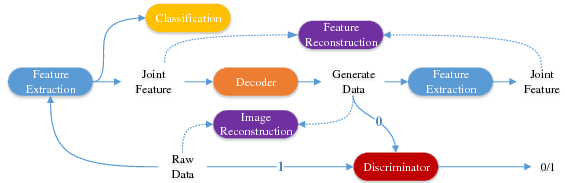}
	\caption{The framework of feature separation. The networks are combined with four modules: the feature extraction module as given in the previous section; a classification module estimating the environmental conditions; a decoder module mapping the extracted feature back to the data domain; a discriminator module distinguishing the generated data and raw data; and two reconstruction loss modules on data and feature domain respectively.}
	\label{fig:feature_separation}
\end{figure*}

\subsubsection{Conditional Entropy Reduction}
$H(z|x)$ measures the uncertainty of feature $z$ given the data sample $x$. The conditional entropy $H(z|x)=0$ can be achieved, if and only if $z$ is the deterministic mapping of $x$.
Thus, reducing $H_{p_{\theta}}(z|x)$ can improve the uniqueness mapping from $x$ to $z$, where $p_{\theta}$ is the parameter in the encoder module. 
However, improving the condition entropy $H_{p_{\theta}}(z|x)$ is intractable, since we can not access the data-label pair $(x,z)$ directly.
An alternative approach is to optimize the upper bound of $H_{p_{\theta}}(z|x)$, and the upper bound can be obtained through the following equation,
\begin{align}
\min_{\theta ,\phi}H_{p_{\theta}}(z|x) &\triangleq \min_{\theta ,\phi} -\sum p_{\theta}(z|x)\log(p_{\theta}(z|x)) \nonumber \\
&=\min_{\theta ,\phi} -\sum p_{\theta}(z|x)[\log(q_{\phi}(z|x))]   \nonumber \\
&    -\sum p_{\theta}(z|x)[\log(p_{\theta}(z|x))-\log(q_{\phi}(z|x))]  \nonumber \\
&=\min_{\theta ,\phi}  H_{p_{\theta}(z|x)}[\log(q_{\phi}(z|x))] \nonumber \\
&    -E_{p_{\theta}(z|x)}[\mathbf{KL}(p_{\theta}(z|x)\|(q_{\phi}(z|x)))], \label{ep:entropy_reduction}
\end{align}
where $\mathbf{KL}$ is the Kullback-Leibler divergence.
And $H_{p_{\theta}}(\log(q_{\phi}(z|x)))$ measures the uncertainty of the predicted feature with a given sample data $x$.
Since we can not extract features from the $q_{\theta}$ directly, we add an additional feature encoder module after the decoder module (see Fig~.\ref{fig:feature_separation}). Eq.~\ref{ep:entropy_reduction} can be converted into
\begin{align}
\min_{\theta ,\phi}H(z|x) &\leq \min_{\theta ,\phi} H_{p_{\theta}(z|x)}[\log(q_{\phi}(z|x))] \nonumber \\
&\triangleq \min_{\theta ,\phi} H_{\hat{z}\sim p_{\theta}(z|x), \hat{x}\sim q_{\phi}(x|\hat{z})}[\log(p_{\theta}(z=\hat{z}|\hat{x}))] \nonumber \\
&=\mathcal{L}_{Feature}(z, \hat{z}), \label{ep:loss_feature}
\end{align}
where $\mathcal{L}_{Feature}$ is the \textit{Feature Reconstruction Loss} between feature extracted from the raw data and the reconstructed data.
As we can see in Eq.~\ref{ep:loss_feature}, the original $H_{p_{\theta}}(z|x)$ is transformed into its upper bound $\mathcal{L}_{Feature}(z, \hat{z})$, 
and the upper bound is reduced  only when the feature domain and data domain are perfectly matched.   

\subsubsection{Feature Extraction Improvement}
Condition entropy reduction sub-module can restrict the mapping uncertainty from data domain to the feature domain, this restriction is highly related to the generalization ability of the encoder module.
For the place recognition task, there will be highly diverse scenes in practice, however, we can only generate limited samples for network training.
In theory, the GAN method use a decoder and discriminator module can learn the the potential feature-to-data transformation with limited samples.
Thus, we improve the data generalization ability by applying GANs.
\begin{align}
\mathcal{L}_{GAN} 
= \min_{\phi}\max_{\omega}E(\log(D_{\omega}(x)) +\label{ep:loss_gan} \\
E_{x\sim q_{\phi}(x|z)}(\log(1-D_{\omega}(x))).\nonumber
\end{align}
As demonstrated by Goodfellow \emph{et.al}~\cite{CNN:GAN}, with iterative updating of the decoder module $q_{\phi}$ and the discriminator module $D_{\omega}$, GAN could pull the generated data distribution closer to the real data distribution, and improve the generalization ability of the decoder module $q_{\phi}$.

\subsubsection{Mutual entropy reduction}

$I(z_{G};z_{C}|x)$ is the mutual entropy, which can be extended by
\begin{align}
I(z_{G};z_{C}|x)=
& H(z_{G}|x)+H(z_{C}|x) -H(z_{G},z_{C}|x),\label{ep11}
\end{align}
where, reducing the mutual entropy is equivalent to reducing the right-hand term in the above equation. 
Since the conditional entropy satisfies $0 \leq H(z_{G}, z_{C}|x)$ , we can find the upper bound of $I(z_{G};z_{C}|x)$ by ignoring  $H(z_{G}, z_{C}|x)$
\begin{align}
\min_{\theta}I(z_{G};z_{C}|x)&\leq \min_{\theta}(H(z_{G}|x)+H(z_{C}|x)). \label{ep:mutual}
\end{align}
 
For the condition-related features, we apply a soft-max based classification module $\mathcal{L}_{Cond}$, to reduce the conditional entropy $H(z_{C}|x)$.
Furthermore, we apply an $L_2$ image reconstruction loss to further restrict the uncertainty $H(z_{G}|x)$ given a sample data $x$,
\begin{align}
\mathcal{L}_{Image} = \|x_{raw}-x_{reco}\|, \label{ep:loss_img}
\end{align}
where $x_{raw}$ and $x_{reco}$ are the raw image and reconstructed one respectively.

By combining Eq.[\ref{ep:loss_feature}, \ref{ep:loss_gan}, \ref{ep:loss_img}] and $\mathcal{L}_{Cond}$, the joint loss function can be obtained as
\begin{align}
\mathcal{L}_{Joint} = &\mathcal{L}_{Feature} + \mathcal{L}_{GAN} + \mathcal{L}_{Cond}+
\mathcal{L}_{Image}. \label{ep:joint} 
\end{align}


\section{Experiment Results}
\label{sec:experiment_results}
In this section, we analyze the performance of our method on two datasets and compare it with three feature extraction methods for the visual place recognition task. 
The experiments are conducted on a single NVIDIA 1080Ti card with 64G RAM on the Ubuntu 14.04 system.
For our method, our can inference the local place image with just $30ms$, and each feature is just $1 kb$.

\begin{figure}
	\centering
	\begin{subfigure}[b]{\linewidth}
		\includegraphics[width=\linewidth]{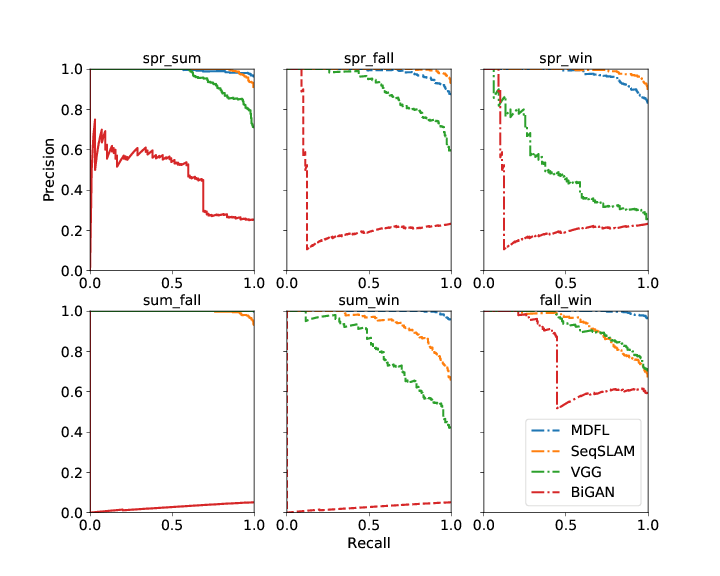}
		\caption{\textit{Nordland} datasets}
		\label{fig:7a}
	\end{subfigure}
	~
	\begin{subfigure}[b]{\linewidth}
		\includegraphics[width=\linewidth]{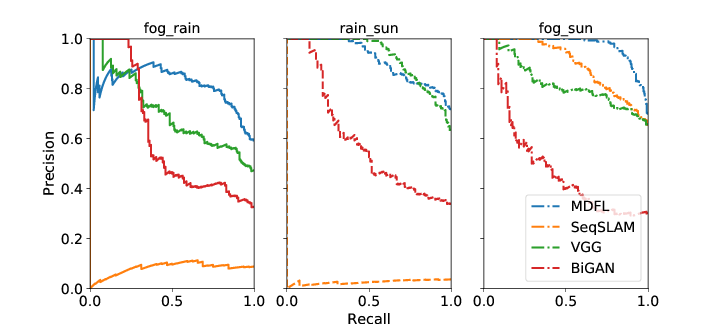}
		\caption{\textit{GTAV} datasets}
		\label{fig:7b}
	\end{subfigure}
	\caption{Precision-Recall curve of various VPR methods on the two datasets. The method is considered to be good if the curve is in the upper-right corner. As we see, MDFL outperforms other methods in most of the cases.}
	\label{Precision-Recall}
\end{figure}

\subsection{Datasets}
The datasets we used here are the \textit{Nordland} dataset \cite{DATASET:NORDLAND} and the \textit{GTAV} dataset \cite{MY:yin2017condition}. 
The \textit{Nordland} dataset was recorded on a train in Norway during four different seasons, and each sequence follows the same track. 
In each sequence, we generate $17885$ frames from the video at $12$ Hz, and the first $16885$ frames of each sequence is used for training, and the last $1000$ frames for testing.  Note that we train on all four \textit{Norldand} seasonal datasets, using the seasonal labels to find the condition dependent/invariant features, and then test on the last 1000 frames of each dataset.
In the training procedure, we randomly select frames and their corresponding status labels from the four sequences.

The second dataset \textit{GTAV}~\cite{MY:yin2017condition} contains trajectories on the same track under three different weather status (sunny, rainy and foggy).
This dataset is more challenging than the \textit{Nordland} dataset, since the viewpoints are variant in the \textit{GTAV} dataset. 
We generate more than $10,000$ frames in each sequence, $9000$ frames are used as training data, and the remaining $1000$ frames for testing. 

For each dataset, all images are resized to $64\times64\times3$ in RGB format. 
The loop closure detection mechanism is followed as in the original SeqSLAM method; sequences of image features are matched instead of a single image. For more details about the structure of the SeqSLAM, we refer the reader to~\cite{VPR:SeqSLAM}.

\subsection{Accuracy Analysis}
To investigate the place recognition accuracy, we compare our feature extraction method with three methods in sequential matching: 
the original feature in SeqSLAM that uses sum of absolute difference as local place feature description; 
convolution layer feature from VGG network, which is trained on the large-scale image classification dataset~\cite{CNN:IMAGENET}; 
adversarial feature learning-based unsupervised feature obtained from the generative adversarial networks~\cite{CNN:BIGAN};
The place is considered as being matched when the distance between current frame and target frame is limited within 10 frames.
We evaluate the performances in the precision-recall curve (PR-curve), area under curve (AUC) index, inference time, and storage requirement.

\begin{figure}
	\centering
	\begin{subfigure}[b]{0.9\linewidth}
		\includegraphics[width=\linewidth]{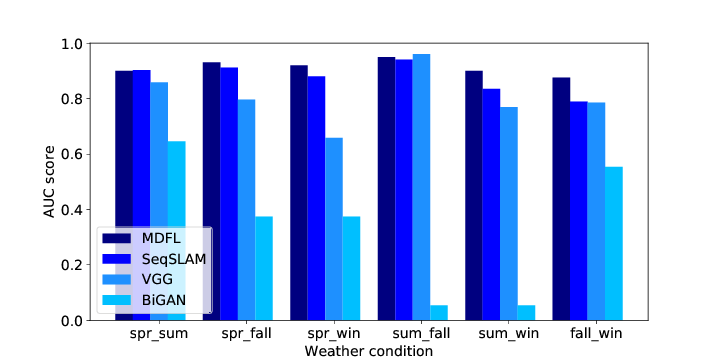}
		\caption{AUC on \textit{Nordland} datasets}
		\label{fig:8a}
	\end{subfigure}
	~
	\begin{subfigure}[b]{0.9\linewidth}
		\includegraphics[width=\linewidth]{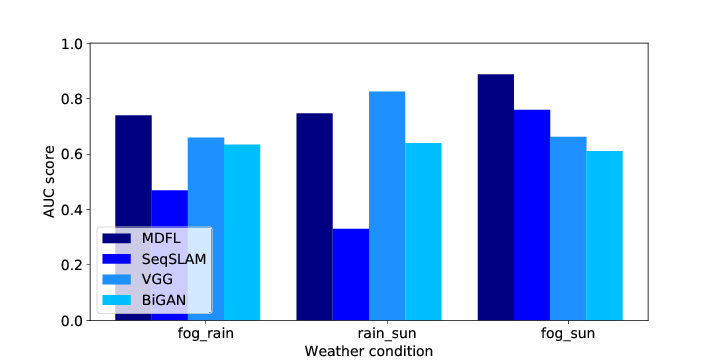}
		\caption{AUC on \textit{GTAV} datasets}
		\label{fig:8b}
	\end{subfigure}
	\caption{AUC index of the various VPR methods. The methods match images from the two datasets under different conditions. }
	\label{AUCindexHist}
\end{figure}

Fig.\ref{Precision-Recall} and ~\ref{AUCindexHist} show the precision-recall curve and AUC index respectively for all the methods on the \textit{Nordland} datasets and the \textit{GTAV} datasets. 
In Fig.~\ref{AUCindexHist}, the label spr-sum, spr-fall, etc. refer to the performance using the same network and same model, with different  testing sequences.

In general, all the methods perform better in the \textit{Nordland} dataset than in the \textit{GTAV} dataset, since the viewpoints are stable and the geometric changes are smooth due to the constant speed of the train. 
In contrast, test sequences in \textit{GTAV} datasets have significant viewpoints differences. Furthermore, limited field of view and multiple dynamic objects in \textit{GTAV} also introduces additional feature noises, which causes a significant difference in scene appearance.

VGG features perform well under normal conditions, such as summer-winter in \textit{Nordland}, but perform poorly under unusual conditions, which indicates that the VGG features trained on normal environmental conditions do not generalize.
BiGAN does not perform well on either datasets, and this is mainly because it does not take into account the condition of the scene and considers all the images as a joint manifold. For example, the same place under different weather conditions will be encoded differently using BiGAN.
Since SeqSLAM uses gray images to ignore the appearance changes under different environmental conditions, the image based features in SeqSLAM are robust against changing conditions as we can see in the \textit{Nordland} dataset.
However, its matching accuracy decreases greatly in the \textit{GTAV} dataset, since raw image features are very sensitive to the changing viewpoints.

In general, MDFL outperforms the above features in most cases of the \textit{NORDLAND} and \textit{GTAV} datasets, and can handle complex situation well, 
but is not the best in some situations, such as spring-summer in \textit{NORDLAND} and rain-summer in \textit{GTAV}.
One potential reason is that, in each dataset, we only consider one type of environmental condition (Season or weather), but we did not take into account the illumination changes. Since the illumination changes continuously, it is not easy to set this type of environmental condition in training.
The geometric features instructed by the environment labels can capture more common geometry details among the multiple weather or season conditions. 
Another advantage lies in the structure of the CapsuleNet-like architecture that enables MDFL to cluster lower level geometry features into high level descriptions.
The benefit of this mechanism can be better seen in the \textit{GTAV} dataset, where the extracted features are more robust to the viewpoints difference. 
Table~\ref{table1} shows the average AUC results of different methods in both datasets, and our MDFL method outperforms all the other methods.

\begin{table}[h]
	\renewcommand{\arraystretch}{1.}
	\caption[m1]{Average AUC index on \textit{Nordland} and \textit{GTAV} datasets}
	\label{table1}
	\centering
	\begin{center}
		\begin{tabular}{l*{5}{c}r} \toprule
			Dataset  & Caps           & SeqSLAM & VGG16 & BiGAN  \\ \midrule
			\textit{GTAV}     & \textbf{0.790} & 0.518   & 0.715 & 0.627  \\ 
			\textit{Nordland} & \textbf{0.912} & 0.876   & 0.804 & 0.345  \\ \bottomrule
		\end{tabular}
	\end{center}
\end{table}

\section{Conclusion} 
\label{sec:conclusion}
In this paper, we propose a novel multi-domain feature learning method for visual place recognition task. 
At the core of our framework lies the idea of extracting condition-invariant features for place recognition under various environmental conditions.
We use a CapsuleNet-based module to capture multi-domain features from the raw image, and apply a feature separation module to indirectly separate condition-related and condition-invariant features.
Based on the extracted condition-invariant features, experiments on the multi-season condition \textit{NORDLAND} dataset and the multi-weather condition \textit{GTAV} datasets, demonstrate the robustness of our method.
The major limitation for our method is that the shallow layer CapsuleNet-based module can only cluster lower level features, and can not capture the semantic descriptions for the place recognition. 
In our future work, we will investigate hierarchical CapsuleNet network module to extract higher level semantic features for place recognition.

\bibliographystyle{plainnat}
\bibliography{main}
\end{document}